\documentclass[10pt,journal]{IEEEtranTCOM}
\usepackage{graphicx} % Required for inserting images
\usepackage[english]{babel}
\usepackage{amsthm}
\usepackage{amsmath}
\usepackage{amsfonts}
\usepackage{mathtools}
\usepackage{booktabs}
\usepackage{url}
\usepackage{geometry}
\usepackage{amssymb}

\newcommand{\be}{\begin{equation}}
\newcommand{\ee}{\end{equation}}

\usepackage[hidelinks]{hyperref}

\usepackage{amsmath}
\usepackage{tikz}
\title{Improving KAN with CDF normalization to quantiles}
\author{\IEEEauthorblockN{Jakub Strawa}\\
\IEEEauthorblockA{StrawAI, Krakow, Poland, jakubstrawaai@gmail.com}\\
\IEEEauthorblockN{Jarek Duda}\\
\IEEEauthorblockA{Faculty of Mathematics and Computer Science, Jagiellonian University, Krakow, Poland, \emph{dudajar@gmail.com}}}
\date{20 July 2025}

\begin{document}
\maketitle
\begin{abstract}
Data normalization is crucial in machine learning, usually performed by subtracting the mean and dividing by standard deviation, or by rescaling to a fixed range. In copula theory~\cite{copula}, popular in finance, there is used normalization to approximately quantiles by transforming $x\to \textrm{CDF}(x)$ with estimated CDF/EDF (cumulative/empirical distribution function) to nearly uniform distribution in $[0,1]$, allowing for simpler representations which are less likely to overfit. It seems nearly unknown in machine learning, therefore, as proposed in \cite{hcr2}, we would like to present some its advantages on example of recently popular Kolmogorov–Arnold Networks (KANs), improving predictions from Legendre-KAN~\cite{chenlegendre} by just switching rescaling to CDF normalization. Additionally, in HCR interpretation, weights of such neurons are mixed moments providing local joint distribution models, allow to propagate also probability distributions, and change propagation direction.

%Kolmogorov–Arnold Networks (KANs) have recently emerged as a compelling alternative to MLPs for universal function approximation, but their training dynamics remain sensitive to input scaling and the choice of polynomial basis. We introduce a Quantile-Normalised KAN, a simple yet powerful refinement that couples (i) Gaussian-quantile normalisation—a per-feature transformation that maps inputs to uniform [0,1] space via the cumulative distribution function (CDF) of the empirical normalised data—and (ii) a scaled Legendre polynomial basis that is orthonormal on 
%[0,1]. The CDF step eliminates distributional skew and heterogeneous variance without requiring batch statistics at inference, while the Legendre basis preserves energy across degrees, yielding numerically stable coefficients.

\end{abstract}
 
\section{Introduction}
Data normalization is very useful for various types of analysis, for example, through batch normalization in neural networks~\cite{batch}. Machine learning uses mainly standard for Gaussian distribution: subtraction of mean and division by standard deviation. Alternatively, for example for combination with Legendre polynomials, which are orthonormal with uniform weight for a fixed range, there was used normalization by rescaling~\cite{chenlegendre}.

However, especially in copula theory~\cite{copula} popular in finance, there is used normalization to nearly uniform distribution in $[0,1]$ by transforming through CDF/EDF like in Fig. \ref{norm}, originally to use various mainly 1 parameter families of copulas. Modeling density as a linear combinations, conveniently in orthornormal basis e.g. of Legendre polynomials, offer practical high parameter descriptions as HCR (hierarchical correlation reconstruction~\cite{hcr1,hcr2}) - Fig. \ref{normex} shows examples of such degree 4 polynomials as density of sample in $[0,1]^2$ normalized by rescaling versus CDF, with the latter offering more uniform density, which is more appropriate to describe with polynomials, especially low order which are less likely to overfit. 

Building neural networks from such neurons containing local joint distribution models, using conditional expected values and only pairwise dependencies, we can simplify to popular Kolmogorov-Arnold Networks (KAN)~\cite{liu2024kan}, additionally: offering  interpretations with mixed moments, joint distribution models, and new capabilities e.g. to estimate mutual information, change propagation direction, or to propagate probability distribution. Therefore, we compare with recent article also using Legendre polynomials in KAN, but with rescale as normalization~\cite{chenlegendre} - we show improvements by replacing it with CDF normalization. Intuitively, thanks to working on more uniform variables, we can improve description with low degree polynomials, which are better to generalize.

\section{CDF/EDF normalization to quantiles}
To normalize variables to nearly uniform distribution in $[0,1]$, we can transform $x\to \textrm{CDF}(x)$ by CDF (cumulative distribution function) like in Fig. \ref{norm}. To find this CDF, we can assume some parametric family like Gaussian, and estimate its parameters e.g. from data sample, batch. In practice we can perform such normalization by subtracting mean from the sample, then divide by its standard deviation like in batch normalization~\cite{batch}, but then additionally transform through normalized Gaussian CDF: 

\begin{figure}[t!]
    \centering
        \includegraphics{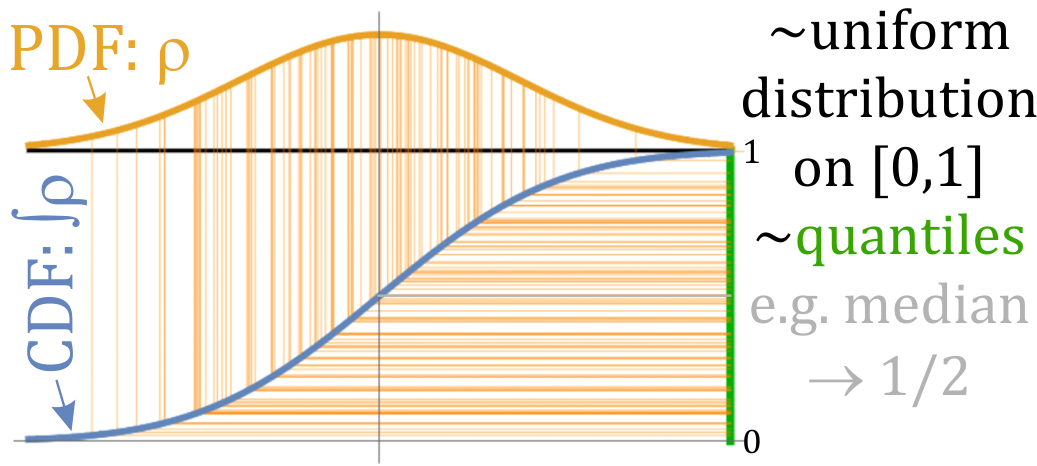}
        \caption{Example of \textbf{CDF normalization} of sample with $x\to\textrm{CDF}(x)$, here from Gaussian distribution. It leads to nearly uniform distribution on $[0,1]$, values are transformed into their approximated quantiles, e.g. median becomes $\approx 1/2$. In practice it is performed by CDF of estimated parameters e.g. by popular and perfect for Gaussian: subtract the mean, divide by standard deviation, but then additionally transform by $\textrm{CDF}_{\mathcal{N}(0,1)}(x)=(1+\textrm{erf}(x/\sqrt{2}))/2$ for normalized Gaussian. Alternatively, we can use EDF: sort the values and transform $i$-th in order into $(i-1/2)/n\in (0,1)$ for size $n$ batch/sample.}
        \label{norm}
\end{figure}

\begin{figure}[t!]
    \centering
        \includegraphics{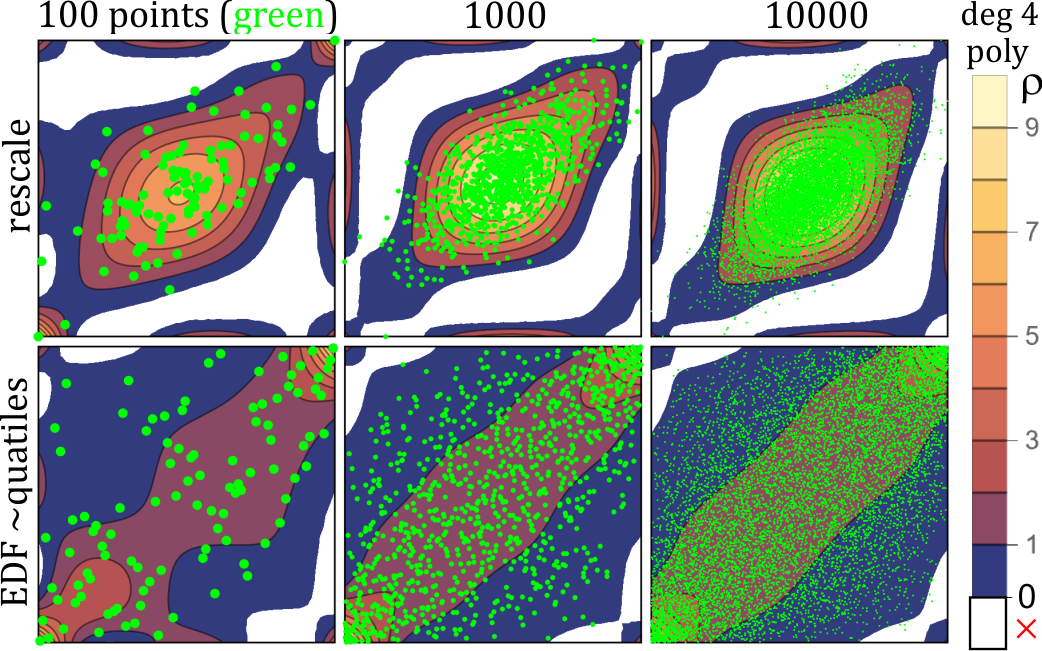}
        \caption{\textbf{Rescale (top) vs CDF (bottom) normalization} on example of generating 100, 1000, 10000 random points (green) from multivariate Gaussian distribution (for covariance matrix $((3,2),(2,3))$), and normalizing separately both coordinates by rescaling to $[0,1]$ (top) or EDF (bottom). Colors are isolines of MSE fitted polynomials as density (HCR) - while they integrate to 1 in the $[0,1]^2$ considered region, they can go below 0 (white areas) - what is not allowed for density, requiring caution for interpretation as density, or calibration to nonnegative. We can see CDF/EDF normalization leads to distribution closer to uniform on $[0,1]^2$, allowing for better density representation by low order polynomials, smaller problematic areas of negative density.  }
        \label{normex}
\end{figure}

\be \textrm{CDF}_{\mathcal{N}(0,1)}(x)=\frac{1}{2}\left(1+\textrm{erf}\left(\frac{x}{\sqrt{2}}\right)\right)\label{cdf0}\ee

Alternatively, especially for non-standard distributions, we can use EDF (empirical distribution function): find order in the sample, and assign $i$-th value in this order to $(i-1/2)/n$ value for size $n$ sample. It can be done in batches, or be trained as interpolated function. For identical values we can their central position to all of them.

However, infinitesimal changes usually maintain the order, leading to zero derivatives in EDF, what is problematic for gradient descent training. To resolve this issue, we can use the fact that derivative of CDF should be the density, which can be estimated e.g. by kernel density estimation from the data sample.

The current tests are using batch normalization followed by normalized Gaussian $x\to (1+\textrm{erf}(x/\sqrt{2}))/2$. 

\section{Kolmogorov-Arnold Theorem and networks}

The \emph{Kolmogorov--Arnold Theorem}~(\cite{kolmogorov,kolm}), also known as the \emph{Superposition Theorem}
or the \emph{Kolmogorov--Arnold Representation Theorem}, is a fundamental result
in approximation theory.  
It states that any continuous multivariate function on a bounded domain
can be represented as a superposition (composition) of a limited number of
one-variable (univariate) functions together with a set of linear operations.

\medskip
Formally, for a continuous function
\[
  f : [0,1]^{d_{\text{in}}} \;\longrightarrow\; \mathbb{R}^{d_{\text{out}}}
\]
defined on the \(d_{\text{in}}\)-dimensional unit hypercube
\([0,1]^{d_{\text{in}}}\), the Kolmogorov--Arnold Theorem guarantees the existence
of continuous univariate functions \(g_q\) and \(\psi_{p,q}\) such that

\begin{figure}[t!]
    \centering
        \includegraphics[width=9cm]{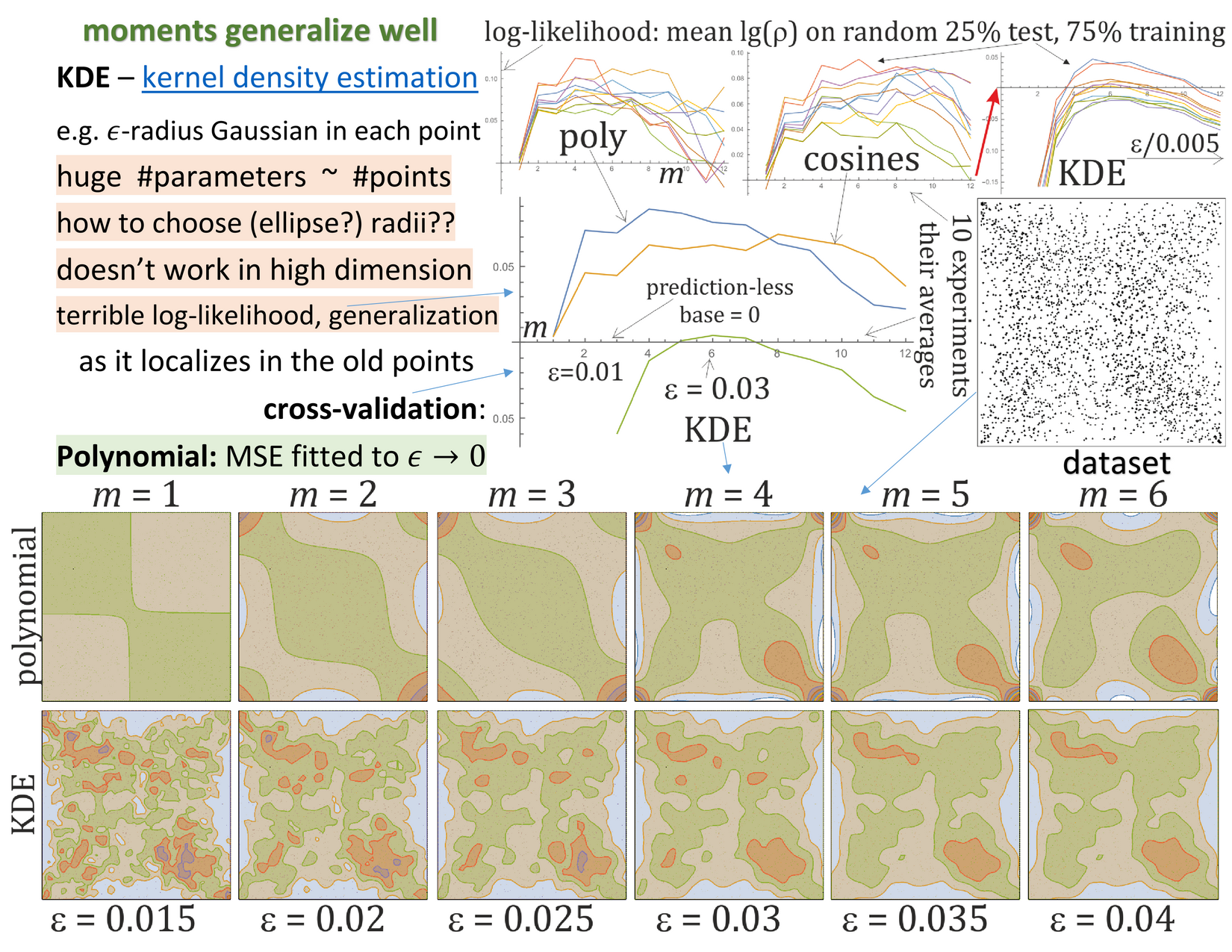}
        \caption{\textbf{Generalization} of \textbf{global} (HCR) vs \textbf{local} (KDE) \textbf{basis} on example of the shown dataset (log-returns for succeeding days CDF normalized to $[0,1]^2$). There were used various size $m$ bases of polynomials and cosines, also KDE (kernel density estimation) with Gaussian kernel of varying width $\varepsilon$. While in cross-validation KDE was barely able to exceed 0 log-likelihood (obtained also by trivial $\rho=1$ assumption), both global bases were able get much higher, the best for $m=4$ degree polynomials: considering up to kurtosis. Better generalization comes from extraction of general features: mixed moments, while KDE just assumes that new points will be close to the old points. }
        \label{global}
\end{figure}

\begin{figure}[t!]
    \centering
        \includegraphics[width=9cm]{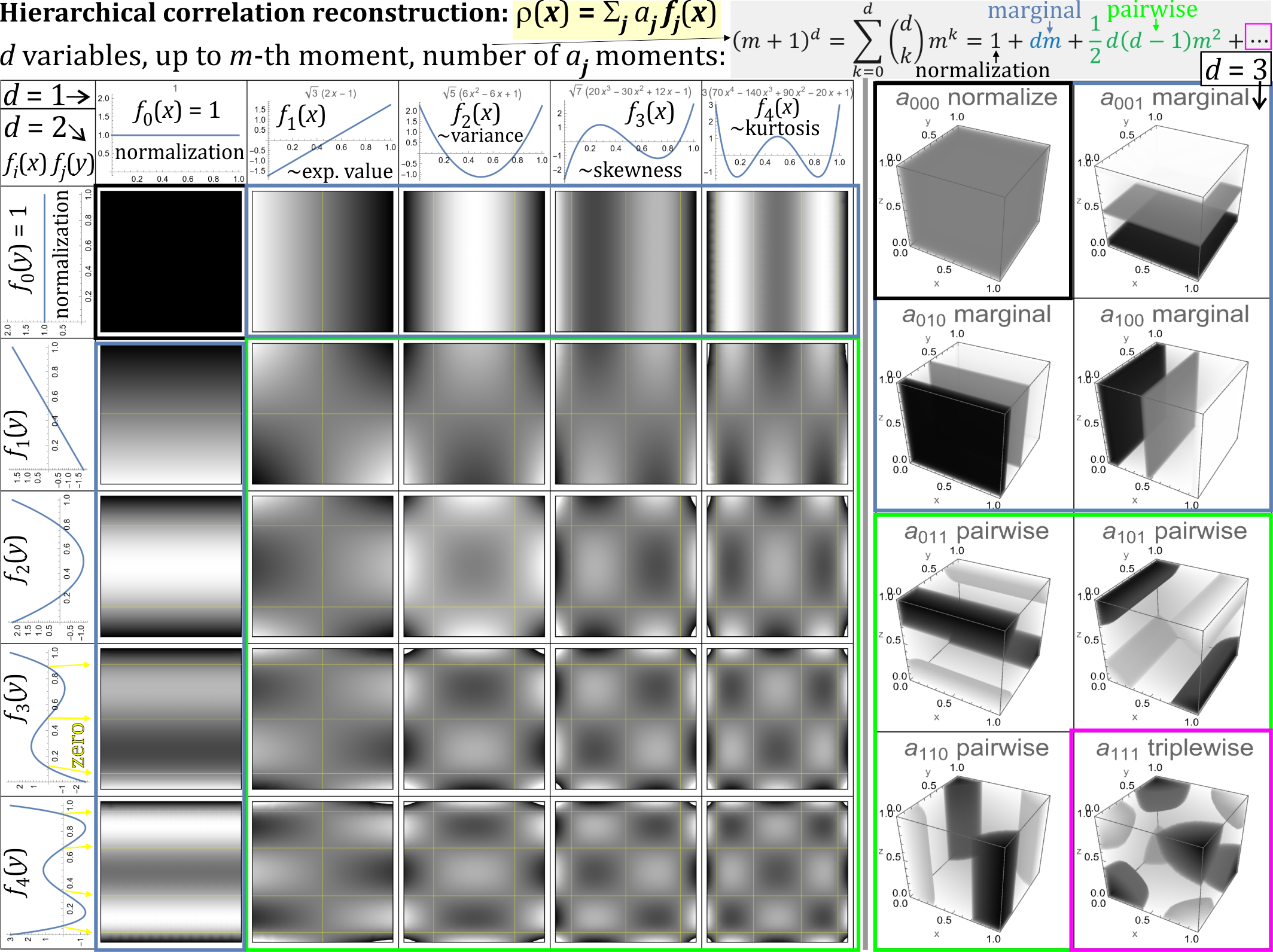}
        \caption{First functions from \textbf{orthonormal polynomial basis} on $[0,1]$ and its product basis $f_{j_1}(x_1)\cdot \ldots\cdot f_{j_d}(x_d)$ for $d=2,3$. Normalizing each variable to nearly uniform in $[0,1]$, for independent variables we would get $\rho=1$ density. In HCR we use $\rho(\textbf{x})=\sum_{\textbf{j}} a_{\textbf{j}}\, f_{j_1}(x_1)\cdot\ldots\cdot f_{j_d}(x_d)$ parametrization of density, allowing to add dependencies in hierarchical way: $\textbf{j}=0$ corresponds to normalization: $a_{00..0}=1$, then $a_\textbf{j}$ with single nonzero index describe marginal distributions, $a_\textbf{j}$ with two nonzero indexes describe pairwise dependencies, and so on. KAN can be viewed as using only pairwise dependencies. }
        \label{basis}
\end{figure}

\begin{equation}
  f(\mathbf{x})
    \;=\;
  \sum_{q = 0}^{2d_{\text{in}}}
    g_q\!\Bigl(
      \sum_{p = 1}^{d_{\text{in}}} \psi_{p,q}(x_p)
    \Bigr),
  \label{eq:kolmogorov-arnold}
\end{equation}
where \(\mathbf{x} = (x_1, x_2, \dots, x_{d_{\text{in}}})\).\\

This theorem has started development of KAN as neural networks based on them, originally using local basis of B-splines~\cite{liu2024kan}. However, global basis like polynomials often has better generalization like shown in Fig. \ref{global}, hence, they are considered as replacement in KAN, e.g. in \cite{ss2024chebyshev,seydi2024exploring,chenlegendre}. Therefore, we will focus on improving the last one, by replacing rescaling with CDF normalization, also introducing to additional HCR interpretations/extensions.

\section{HCR interpretations/extensions}
\begin{figure}[t!]
    \centering
        \includegraphics[width=9cm]{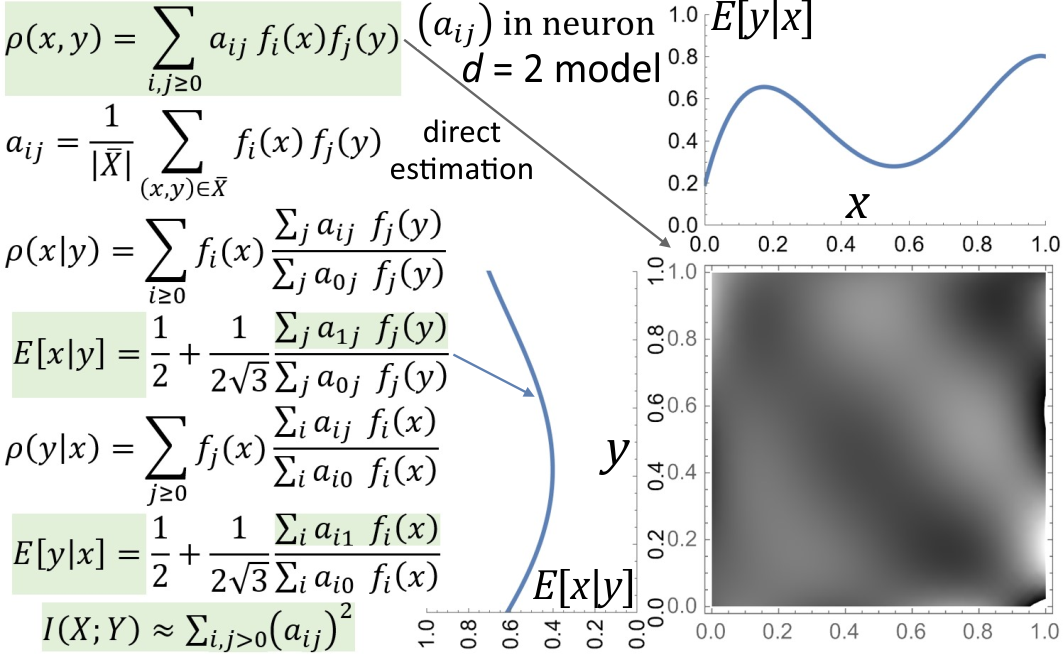}
        \caption{2D example of HCR density parametrization, its direct estimation, conditional densities and expected values, also approximate formula for mutual information. HCR neuron~\cite{hcr2} containing such $(a_{ij})$ parameters describing local joint distribution can propagate in any direction, values or probability distributions. Increasing the number of connections but restricting to pairwise dependencies, for value propagation such neural network is equivalent to KAN.}
        \label{2d}
\end{figure}

We will now briefly introduce to HCR (hierarchical correlation reconstruction)~(\cite{hcr1,hcr2}), which offers interpretation of e.g. polynomial-based KANs: for coefficients as mixed moments (contributing to entropy/mutual information), entire neurons as models of local joint distribution. It also allows to extend KAN for additional capabilities, like including higher local dependencies, propagation of probability distributions, or changing propagation directions, like Fig. \ref{2d}.

\textbf{HCR approach models joint density as a linear combination} - conveniently on $[0,1]^d$ for $d$ variables normalized to nearly uniform in $[0,1]$ (CDF), e.g. using product basis: 
\be \rho(\textbf{x})=\sum_{\textbf{j}\in B} a_{\textbf{j}}\, f_{j_1}(x_1)\cdot ...\cdot f_{j_d}(x_d)\label{rho}\ee
for $\textbf{x}=(x_1..x_d)$, $\textbf{j}=(j_1..j_d)$ and $B$ some considered basis. 

While such density parametrization is very convenient, and integration to 1 (normalization) is obtained by just $a_{00..0}=1$, it has issue of sometimes getting negative density, like white regions in Fig. \ref{normex}. In standard HCR applications such density is calibrated by using e.g. $\max(\rho,0.1)$ and normalizing to integrate to 1. However, this integration is numerically costly. For many applications like neural networks this issue seems practically negligible, so we will not repair $\rho<0$ here, however, we should have in mind potential consequences of such approximation.

Assuming \textbf{orthogonal basis} in $[0,1]$: 
\be \int_0^1 f_j(x) f_k(x) dx =\delta_{jk}\ee
we can directly and independently MSE estimate~\cite{HCR0} coefficients from data sample $X$ (used for Fig. \ref{normex}, \ref{global}):
\be a_{\textbf{j}} = \frac{1}{|X|} \sum_{\textbf{x}\in X} f_{j_1}(x_1)\cdot ...\cdot f_{j_d}(x_d) \ee

We will focus on polynomial basis, but different ones are also worth considering, especially DCT/DST (discrete cosine/sine transform) for periodic data.

Orthogonal polynomial basis in $[0,1]$ are \textbf{rescaled Legendre polynomials} polynomials, shown in Fig. \ref{basis}, starting with:
\be f_0 = 1\qquad\textrm{corresponds to normalization}\label{leg}\ee
$$ f_1(x)=\sqrt{3} (2 x-1)\qquad \sim\textrm{expected value}$$
$$f_2(x) =\sqrt{5} \left(6 x^2-6 x+1\right) \qquad \sim\textrm{variance} $$
$$ f_3(x) = \sqrt{7} \left(20 x^3-30 x^2+12 x-1\right)\qquad\sim\textrm{skewness}$$
$$ f_4(x) =3 \left(70 x^4-140 x^3+90 x^2-20 x+1\right)\qquad\sim\textrm{kurtosis} $$
Coefficients $a_j$ in this basis are \textbf{moments}, approximately cumulants: $a_1\sim$ expected value, $a_2\sim$ variance, $a_3\sim$ skewness, $a_4\sim$ kurtosis. For $f_{j_1}..f_{j_d}$ product basis, as $f_0=1$, coefficient $a_{\textbf{j}}$ do not depend on variables having zero index $j_i=0$, and describe dependence between corresponding moments for non-zero indexes. For example $a_{3002}$ describes dependence between 3rd moment ($\sim$ skewness) of the 1st variable and 2nd moment ($\sim$ variance) of the 4th variable in $d=4$ dimensions.

\subsection{Conditional density/value and KAN as special case}
Having (\ref{rho}) joint density model $\rho$, we can substitute some variables and normalize ($1$ for $i=0$), getting approximated conditional (joint) densities. For example in $d=3$ we get:
\be \rho(x|y,z)=\sum_i f_i(x) \frac{\sum_{jk} a_{ijk} f_j(y) f_k(z)}{\sum_{jk} a_{0jk} f_j(y) f_k(z)} \ee

To propagate (conditional) values, we can calculate  \textbf{expected value} of such conditional density. The contributions become $\int_0^1 x f_0(x)dx =1/2$, $\int_0^1 x f_1(x)dx =1/\sqrt{12}$, and zero for higher moments, leading to formulas including only $i=0$ normalization and $i=1$ the first moment:
\be E[x|y,z]=\frac{1}{2} +\frac{1}{\sqrt{12}} 
\frac {\sum_{jk} a_{1jk} f_j(y) f_k(z)}{\sum_{jk} a_{0jk} f_j(y) f_k(z)}  \ee

Building neural networks from them, with interlayer normalization which shifts and rescales, we can use only the nominator here: $\sum_{jk} a_{1jk} f_j(y) f_k(z)$. Restricting to only \textbf{pairwise dependencies}: coefficients with two nonzero indexes, it becomes sum of trained 1-variable polynomials as in KAN:
\be {\sum_{jk} a_{1jk} f_j(y) f_k(z)} \xrightarrow[\textrm{KAN-like}]{\textrm{pairwise only}} \sum_j a_{1j0} f_j(y) +\sum_k a_{10k} f_k(z)\label{KANl}
\ee
This way we can interpret e.g. polynomial based KAN as HCR - its coefficients as moments, representing local joint distribution, what seems valuable e.g. for interpretability/explainability.

Also allowing to extend in various ways, for example consciously add triplewise and higher dependencies, propagate probability distributions, or flexibly change propagation direction.

\subsection{Entropy, mutual information evaluation}
Using HCR interpretation of KAN, we can also get information theoretic evaluation useful for interpretability/explainability, or training e.g. by information bottleneck~\cite{information}.

Using $\ln(1+a)\approx a$ approximation and orthogonality, we can approximate \textbf{entropy} in nits ($1/\ln(2)\approx 1.44$ bits) from the parameters:
\be H(X)=-\int_{[0,1]^d}\rho(\mathbf{x})\ln(\rho(\mathbf{x}))d\mathbf{x} \approx -\sum_{\mathbf{j}\in B^+} (a_\mathbf{j})^2   \ee 
for $B^+ = B\backslash \{\textbf{0}\}$ basis without normalization. We can analogously find formula for joint distribution $H(X,Y)$, getting \textbf{mutual information} approximation as just sum of squares of nontrivial coefficients between variables (can be multivariate):
\be I(X;Y)=H(X)+H(Y)-H(X,Y) \approx \sum_{\mathbf{j}_x\in B_{X}^+}\  \sum_{\mathbf{j}_y\in B_{X}^+} \left(a_{(\mathbf{j}_x,\mathbf{j}_y)}\right)^2 \label{mif} \ee 

\section{Legendre-KAN \cite{chenlegendre}}
\subsection{MinMax normalization by rescaling}
\label{sec:minmax}
In the original LegendreKAN paper, the authors proposed MinMax normalization, also called \emph{feature scaling}. It linearly rescales raw values so that every feature lies inside a predetermined range, usually
$[0,1]$. For a given scalar input $x$ the transform is:

\[
\operatorname{MinMax}(x)
  \;=\;
  \frac{x - x_{\min}}{x_{\max} - x_{\min}},
\]
where $x_{\min}$ and $x_{\max}$ denote, respectively, the minimum and
maximum of the feature over the reference set (an individual image, a
mini-batch, or the whole training corpus).

\begin{itemize}
  \item \textbf{Affine invariance.}  The operation is an affine map
        that preserves relative spacing: the smallest value is mapped
        to~0, the largest to~1, and everything else sits proportionally
        in between.
  \item \textbf{Unit-scale inputs.}  By forcing every (rescaled) feature
        to share the same dynamic range, the optimiser can use a single
        global learning rate without being dominated by disproportionately
        large or small coordinates.
  \item \textbf{Limitations.}  The method is sensitive to outliers
        because $x_{\max}$ and $x_{\min}$ are extreme statistics.
        In images such as MNIST, the majority of background pixels end
        up \emph{exactly} at~0, creating degenerate mass points that
        can hamper gradient flow in higher-order polynomial layers (see
        Section~\ref{sec:experiments}). Replacing with CDF normalization repairs this limitation.
\end{itemize}

\subsection{Legendre Polynomials}
Legendre polynomials form a sequence of orthogonal polynomials that arise naturally in potential theory, numerical integration, and many areas of physics (e.g.\ quantum mechanics and electromagnetism).  
They are originally defined on the $[-1,1]$ interval and satisfy the three-term recurrence relation:

\be P_0(x) = 1\qquad \qquad P_1(x) = x \label{eq:recurrence}\ee
$$\textrm{for}\ k\ge 1:\quad (k+1)\,P_{k+1}(x) = (2k+1)\,x\,P_k(x) - k\,P_{k-1}(x) $$

Here as probabilities/quantiles we work in $[0,1]$ range instead, requiring to rescale them for $\int_0^1 f_j(x) f_k(x) dx=\delta_{jk}$ orthonormality: \be f_k(x)=\sqrt{2k-1}\,P_k(2x-1) \ee with the first ones presented in (\ref{leg}) and Fig. \ref{basis}.\\

%\paragraph{First few Legendre polynomials}
%\begin{align}
%P_0(x) &= 1, \tag{4}\\
%P_1(x) &= x, \tag{5}\\
%P_2(x) &= \tfrac12\!\bigl(3x^{2}-1\bigr), \tag{6}\\
%P_3(x) &= \tfrac12\!\bigl(5x^{3}-3x\bigr), \tag{7}\\
%P_4(x) &= \tfrac18\!\bigl(35x^{4}-30x^{2}+3\bigr).\tag{8}
%\end{align}

Key properties that make Legendre polynomials attractive for approximation:
\begin{itemize}
  \item \textbf{Orthogonality.} The family $\{P_n\}$ is orthogonal on $[-1,1]$ with respect to the \emph{constant} weight $w(x)=1$.
    %    \[
    %      \int_{-1}^{1} P_m(x)\,P_n(x)\,dx = \frac{2}{2n+1}\,\delta_{mn}.
    %    \]
        Orthogonality decouples the coefficients of a Legendre expansion, which improves numerical stability and convergence.

  \item \textbf{Gauss–Legendre Quadrature.} The roots of $P_n$ serve as nodes for Gauss–Legendre quadrature, yielding $2n\!+\!1$-point rules that integrate polynomials of degree $\,\le 2n\!+\!1$ exactly—one of the most efficient Gaussian quadrature formulas.

  \item \textbf{Rapid Convergence.} For sufficiently smooth functions the error of a Legendre expansion decays exponentially with the degree, so comparatively low-order polynomials can deliver high accuracy.

  \item \textbf{Recursive Computation.} Equation~\eqref{eq:recurrence} enables fast, stable evaluation of $P_n(x)$ and its derivatives without explicit high-degree polynomial arithmetic—important for spectral and finite-element methods.
\end{itemize}

\subsection{Input Normalization}
\label{sec:gaussian_cdf_norm}

When the inputs are expected to follow an (approximately) Gaussian distribution,  
a convenient way to map them into a bounded interval is to standardize each feature
and then pass it through the CDF of a standard normal distribution.  
Compared with a purely linear rescaling, this transformation is smooth, strictly
monotone, and compresses outliers.

\paragraph{Forward pass.} As in batch normalization~\cite{batch}, given a size $n$ mini-batch %\footnote{%
%$\mathbf X\in\mathbb R^{n\times d_{\mathrm{in}}}$ with rows
%$\mathbf x_i^{\!\top}$, $i=1,\dots,n$.} 
$\mathbf X$, compute the per-feature mean and
(unbiased) standard deviation (with a small constant
$\varepsilon>0$ added for numerical stability, default $10^{-5}$):
% niepotrzebny chaos z tymi wektorami, wystarczy "per-feature" i na pojedynczych

\be 
\mu 
      = \frac1n\sum_{i=1}^{n} x_i,
\qquad
\sigma 
      = \sqrt{\varepsilon + \frac1{n-1}\sum_{i=1}^{n}( x_i-\mu)^2}
\ee

The \textbf{Gaussian + CDF-normalized} tensor is then obtained element-wise via per-feature:
\be
\mathbf U
  = \textrm{CDF}_{\mathcal{N}(0,1)}\Bigl(\frac{\mathbf X-\mu}{\sigma}\Bigr)
  = \tfrac12\!\left(1+\operatorname{erf}
          \Bigl(\tfrac{\mathbf X-\mu}{\sigma\sqrt2}\Bigr)\right),
\ee
%where $\Phi$ is the standard-normal cumulative distribution function.  

\section*{Model Architecture}

\be
\varphi(\mathbf{x}) 
= \sum_{i=1}^{n} \sum_{j=0}^{d} 
w_{ij} \, f_{j}\left(
    \underbrace{
        \tfrac{1}{2} \left( 1 + \operatorname{erf}\left(\frac{x_i}{\sqrt{2}}\right) \right)
    }_{\scriptstyle u_i}
\right)
\ee

The original Legendre KAN uses \textbf{LayerNorm}\footnote{https://docs.pytorch.org/docs/stable/generated/torch.nn.LayerNorm.html} adding $\beta,\gamma$ trainable parameters below, which for further normalized CDF should be $\beta=1,\gamma=0$. For any case it was tested, showing its training indeed do not improve performance - further "learnable" denotes training $\beta,\gamma$, and "FixedNorm" denotes fixing them. The authors in the original paper also use LayerNorm at the end of each layer, we propose the usage of LayerNorm before the layer, which can be seen in section E.

\be\textrm{LayerNorm:}\quad \textrm{LN}(\mathbf X)=\frac{\mathbf X-\textrm{E}[\mathbf X]}{\sqrt{\textrm{Var}[\mathbf X]+\varepsilon}} \beta +\gamma \ee

\subsection{CDFKal Layer}

For an input \(h\in\mathbb R^{n}\) and learnable \(W\in\mathbb R^{m\times n(d+1)}\)
\[
\boxed{%
  \begin{array}{c}
      \underbrace{h}_{\mathbb{R}^{n}}\\[4pt]
      \stackrel{\textrm{LayerNorm (LN), then CDF}}{\downarrow}\\[4pt]
      u_i = \frac{1}{2}(1+\textrm{erf}(\textrm{LN}(h_i/\sqrt{2})))\\[4pt]
      \stackrel{\mathcal{P}_{0:d}\ \textrm{polynomials of }0..d\textrm{ degree}}{\downarrow}\\[4pt]
      p=\bigl[\mathcal{P}_{0:d}(u_{1});\,\dots;\,\mathcal{P}_{0:d}(u_{n})\bigr]\\[4pt]
      \stackrel{W\textrm{ trained weights/coefficients}}{\downarrow}\\[4pt]
      y = W\,p \in \mathbb{R}^{m}
  \end{array}}
\]

Note: Wanting to use SiLU function, like in the original KAN article, one can simply wrap the input $h$ as a residual connection and add it after combining computed Legendre polynomials with their weights.

\subsection{CDFKAN Network}

\[
\begin{aligned}
x_{0}       &= \operatorname{vec}(x)                             \\[0.3em]
\hat x_{0}  &= \operatorname{LN}_{0}(x_{0})                      \\[0.3em]
h_{1}       &= \mathrm{CDFKal}_{1}\!\bigl(\hat x_{0}\bigr)         \\[0.3em]
\hat h_{1}  &= \operatorname{LN}_{1}(h_{1})                      \\[0.3em]
h_{2}       &= \mathrm{CDFKal}_{2}\!\bigl(\hat h_{1}\bigr)         \\[0.3em]
\hat h_{2}  &= \operatorname{LN}_{2}(h_{2})                      \\[0.3em]
\text{logits} &= \mathrm{CDFKal}_{3}\!\bigl(\hat h_{2}\bigr)       \\[0.3em]
\end{aligned}
\]

\section{Experiments and Results}
\label{sec:experiments}
%===========================================================

\subsection{Goal}
We investigate how the choice of feature normalization affects the
generalization ability of Legendre–polynomial-based kernels.
Specifically, we compare the classical \textbf{MinMax} scaling used in the original LegendreKAN implementation, which will be denoted as 
\textbf{KAL\_NET} against a two-step \textbf{Gaussian-CDF}\footnote{%
Each feature is first standardized to zero mean and unit variance, then
mapped through the cumulative distribution function (CDF) of a
\(\mathcal{N}(0,1)\) variable.}
transformation employed by \textbf{CDFKAL\_NET}.  A third variant,
\textbf{CDFKAL\_NET\_FIXEDNORM}, is just the affine-free version of \textbf{CDFKAL\_NET} where 
in the standardization, the parameters are not trained throughout the network.
Last variant, \textbf{CDFKAL\_SILU} is equivalent to \textbf{CDFKAL\_NET} with single SiLU\footnote{https://docs.pytorch.org/docs/stable/generated/torch.nn.SiLU.html} activation added as a residual connection on the input of every layer.

\subsection{Setup}
All the models are trained using the same 3-layer network (not counting the LayerNorm as a layer), the same MNIST dataset and evaluated on a held-out test split.
All code and training scripts will be available in the repository: 
\href{https://github.com/jakubstrawa1/CDFLegendreKan}{\texttt{github.com/jakubstrawa1/CDFLegendreKan}}
We sweep the Legendre polynomial degree \(d \in \{3,\dots,11\}\). Hyper-parameters (learning
rate, batch size) are fixed across the sweep to
isolate the effect of normalization.

\subsection{Networks and Training}

\paragraph{Architectures.}
\begin{itemize}
  \item \textbf{\texttt{KAL\_NET}} —  Original Legendre KAN implementation with per-layer MinMax rescaling, double \texttt{SiLU} non-linearities (on input of they layer as well as the output) and a final \texttt{LayerNorm}. 
  \item \textbf{\texttt{CDFKAL\_NET\_FIXEDNORM}} — CDF variant with a \texttt{LayerNorm} whose scale and shift are \emph{frozen} at $\gamma=1,\ \beta=0$. No \texttt{SiLU}. 
  \item \textbf{\texttt{CDFKAL\_NET}} — identical to the previous model, but $\gamma$ and $\beta$ are trainable. Also without \texttt{SiLU}.
  \item \textbf{\texttt{CDFKAL\_SILU}} — identical to the CDFKAL\_NET, but with single \texttt{SiLU} activation function on the input of every layer. 

\end{itemize}

\paragraph{Training setup.}
All networks are trained for $20$ epochs with Adam (learning rate $10^{-3}$) on a randomly sampled $20\,000$-image subset of the MNIST training set; performance is reported on the standard $10\,000$-image test split. We used the cross‐entropy loss.

\begin{figure}[htbp]
  \centering
  \includegraphics[width=0.9\linewidth]{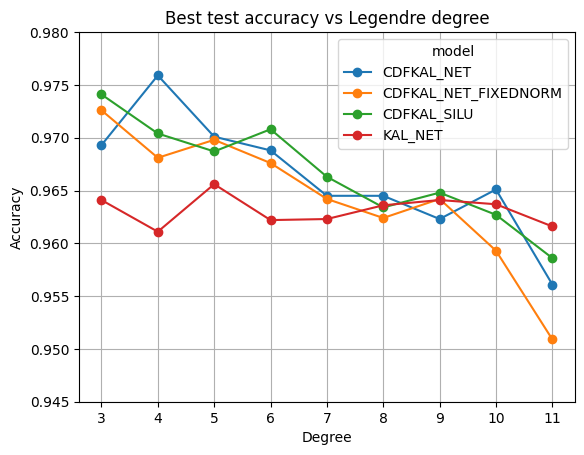}
  \caption{Best test accuracy obtained with all models
           as a function of the Legendre polynomial degree. Observe that low degrees offer the best generalization, for which uniformization with CDF has allowed for better representation.}
  \label{fig:acc_vs_degree}
\end{figure}

\subsection{Results}
\paragraph{Normalization matters.}
Replacing the MinMax scaling used in \textbf{KAL\_NET} with our Gaussian-CDF normalization (z‑score $\rightarrow$ CDF) in \textbf{CDFKAL\_NET} yields a consistent gain of roughly 0.5–2 percentage points in test accuracy across all Legendre degrees lower than 8th degree. (Fig.~\ref{fig:acc_vs_degree}). This distribution‑aware transform more effectively equalizes feature scales, making polynomial features easier to learn. Even the fixed‑norm variant (\textbf{CDFKAL\_NET\_FIXEDNORM}) and the SiLU‑based version (\textbf{CDFKAL\_SILU}) outperform the original MinMax approach, although they tend to compound after polynomial degree greater than 8.

\begin{figure}[htbp]
  \centering
  \includegraphics[width=0.9\linewidth]{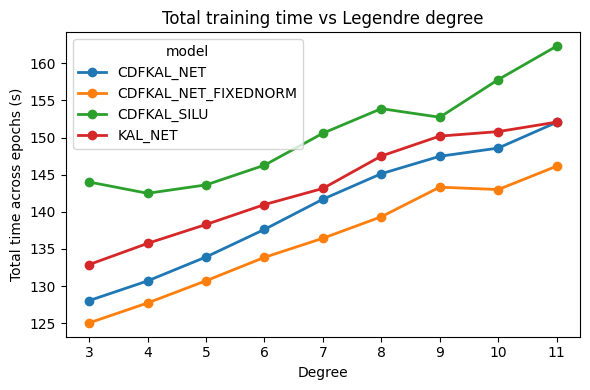}
  \caption{Total wall‑clock training time (seconds) required for each model to
           reach convergence across Legendre polynomial
           degrees.}
  \label{fig:time_vs_degree}
\end{figure}

\paragraph{Training efficiency.}
Beyond accuracy, \textbf{CDFKAL\_NET\_FIXEDNORM} and \textbf{CDFKAL\_NET} also converges noticeably
faster (Fig.~\ref{fig:time_vs_degree}). Eliminating the learnable affine
parameters (Orange line) shaves roughly \(5\text{–}10\,\%\) off the wall‑clock time compared
with the baseline \textbf{KAL\_NET} and more than \(15\,\%\) relative to the
original \textbf{CDFKAL\_NET}. It is also worth mentioning, that although \textbf{CDFKAL\_SILU} is on par with the other 2 CDF variants accurace-wise, it is the slowest in terms of computation.
In practice, the fixed‑norm variant thus offers
the best of both worlds: higher test accuracy and lower computational cost.

For reference, benchmarks were performed on Google Colab's GPU T4.

\begin{figure*}[!t]
  \centering
  \includegraphics[width=\textwidth]{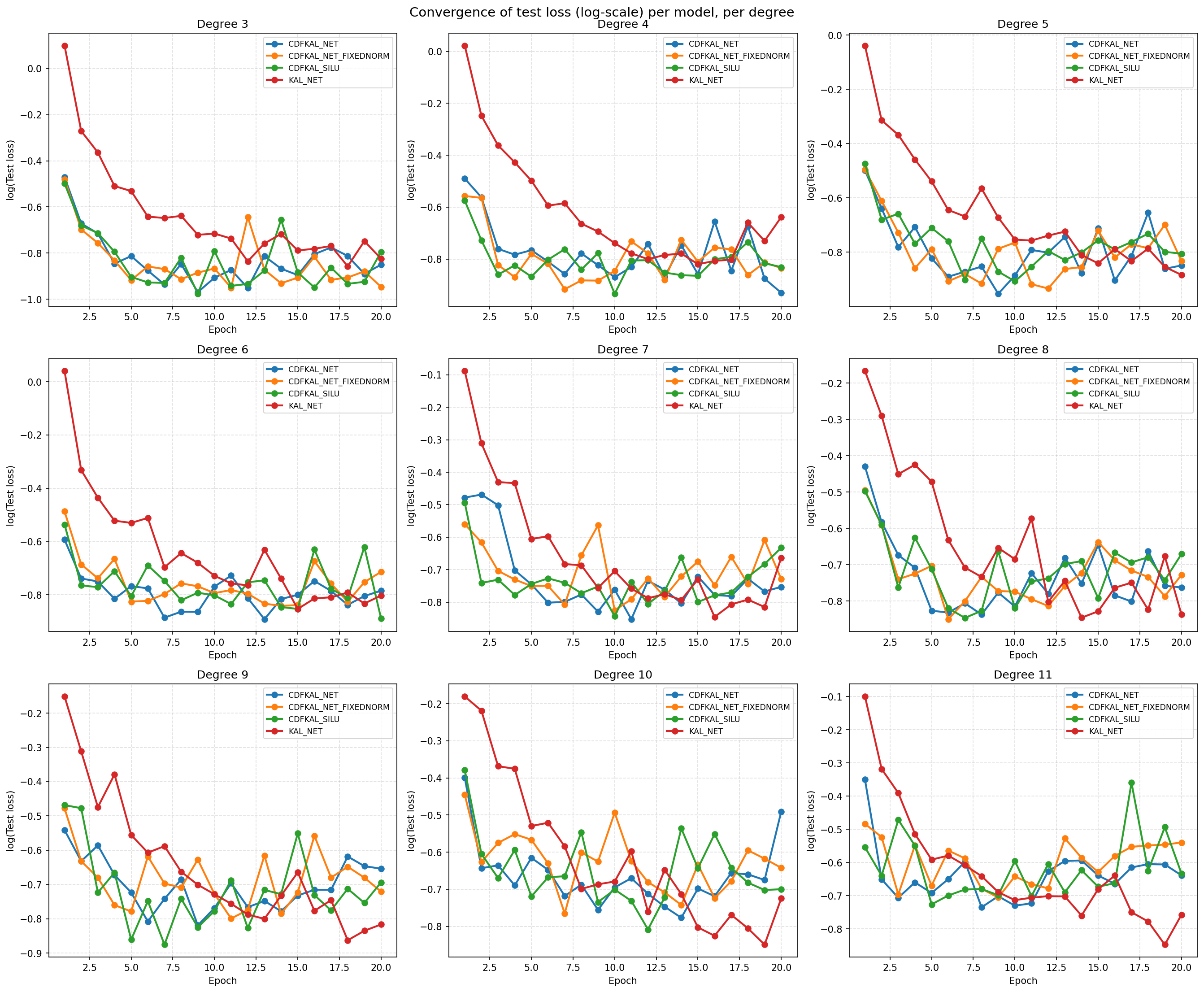}
  \caption{%
    Test‐set convergence of our Legendre KAN variants on polynomial
    regression tasks of degree 3 through 11. Each subplot shows the
    evolution of $\log_{10}(\text{test loss})$ over 20 training epochs for
    four models:
    \textbf{KAL\_NET} (red), the CDF‑normalized variant with frozen
    LayerNorm (\textbf{CDFKAL\_NET\_FIXEDNORM}, orange),
    the CDF‑normalized variant with learnable LayerNorm
    (\textbf{CDFKAL\_NET}, blue), and
    \textbf{CDFKAL\_SILU}, which adds SiLU activation to CDFKAL\_NET
    (green). The relative improvement persists across degrees but
    gradually shrinks on the hardest (highest‐degree) fits.%
  }
  \label{fig:kan_degree_convergence}
\end{figure*}

\paragraph{Convergence across polynomial degrees.}
Figure~\ref{fig:kan_degree_convergence} demonstrates that, without any
MinMax input scaling, our CDF‑normalized KAN models on average converge $\approx 2\times$ faster.
Although all models slow down on higher‐degree polynomials, the CDF‐based approaches still
retain a clear advantage over the original KAN in almost every setting. This shows that our approach can be trained in half as many epochs and still retain the same test accuracy as the MinMax approach.

\begin{figure}[htbp]
  \makebox[\linewidth][l]{%
      \includegraphics[width=\linewidth]{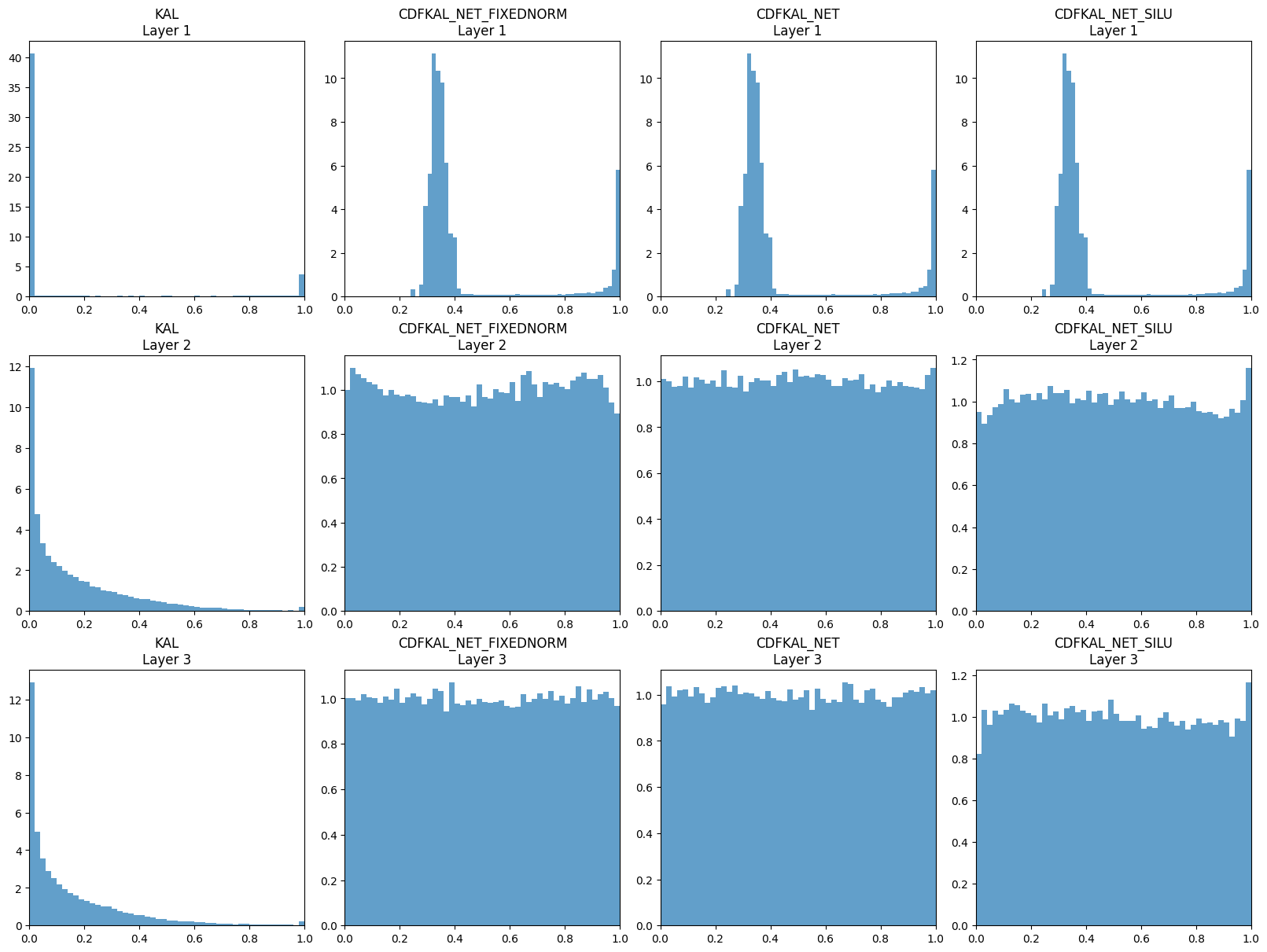}}%
  \caption{%
    Histograms after normalization for 3 layers (rows) using 4 normalization approaches (columns). We can observe the original KAL has very non-uniform densities, what seems the reason for lower accuracy as low degree polynomials cannot represent them well. We can also observe that learnable normalization has similar uniformity, hence can be removed to reduce training cost. The first layer is far from uniform, but essentially more uniform than for KAL - better uniformization could be obtained with EDF.}
  \label{fig:pixel_hists}
\end{figure}

\section{Conclusion}
\label{sec:conclusion}

This work systematically disentangled the influence of feature normalisation
from all other architectural choices in the original Legendre–polynomial‐based
KAN. Main take‑aways emerge:

\begin{enumerate}
  \item \textbf{Gaussian‑CDF normalization wins on low degree polynomials}  Across all low-degree polynomials, one or more members of the
        \textsc{CDFKal} family not only achieves 0.5–2 p.p. higher test
        accuracy on MNIST than the baseline \textsc{Kal\_Net} (Fig. 6),
        but also converges $\approx 2\times$ faster in terms of epochs. By applying
        LayerNorm before the CDF transform, inputs become both variance
       –equalised and support‑bounded, exactly matching the orthonormality
        assumptions of the Legendre basis.
  \item \textbf{Learnable vs.\ frozen LayerNorm.}  Allowing the LayerNorm
        scale and shift to be trained (\textsc{CDFKal\_Net})
         sets the best test accuracy throughout the whole experiment, s
         suggesting that a small amount of adaptive rescaling further refines the input distribution.
  \item \textbf{Single SiLU activation adds little overhead.}  Introducing
        a SiLU nonlinearity immediately after the CDF (\textsc{CDFKal\_SiLU})
        slightly increases training complexity but does not degrade—and in
        some cases even improves—both convergence speed and final accuracy.
    \item \textbf{There is accuracy improvement on the test set to degree 7}, then we can observe drop in the evaluation corresponding to overfitting. Uniformity obtained by CDF normalization has allowed for better representation with such relatively low degree polynomials.

\end{enumerate}

\section{Future work.}
Several directions look promising:
\begin{enumerate}
  \item \textbf{Broader datasets.}  Validate the Gaussian-CDF normalization on higher-resolution and higher-diversity benchmarks—e.g.\ CIFAR-10/100, SVHN, Fashion-MNIST, ImageNet—and on non-vision modalities such as tabular data, speech and text.
  \item \textbf{Robustness scenarios.}  Study the behavior under distribution shift, class imbalance, and adversarial or noisy inputs to see whether the improved coefficient balance carries over to robustness.
  \item \textbf{HCR interpretation/extensions}: Interpretation of the found coefficients as moments, joint distributions for better explainability, information theoretic evaluation, and new capabilities like propagation of probability distribution or changing propagation direction.  
\end{enumerate}

\bibliographystyle{IEEEtran}
\bibliography{cites}
\end{document}